\documentclass{ifacconf}

\usepackage{graphicx}      
\usepackage{natbib}        

\let\theoremstyle\relax
\usepackage{amsmath,amsfonts,amssymb,bm,amsthm} 
\usepackage{xcolor} 
\usepackage{tikz} 
\usetikzlibrary{angles, quotes}
\usepackage{url}
\usepackage{subcaption}

\usepackage{comment}

\newtheorem{remark}{Remark}

\usepackage{todonotes}

\begin{document}

\begin{frontmatter}

\title{Mass-Adaptive Admittance Control for Robotic Manipulators} 

\thanks[footnoteinfo]{Sandia National Laboratories is a multi-mission laboratory managed and operated by National Technology \& Engineering Solutions of Sandia, LLC., a wholly owned subsidiary of Honeywell International, Inc., for the U.S. Department of Energy’s National Nuclear Security Administration under contract DE-NA0003525. This paper describes objective technical results and analysis. Any subjective views or opinions that might be expressed in the paper do not necessarily represent the views of the U.S. Department of Energy or the United States Government.}

\author[First]{Hossein Gholampour} 
\author[Second]{Jonathon E. Slightam} 
\author[First]{Logan E. Beaver}

\address[First]{Department of Mechanical \& Aerospace Engineering, Old Dominion University, Norfolk, VA 23529, USA (e-mail: \{mghol004, lbeaver\}@odu.edu).}
\address[Second]{Unmanned Systems \& Autonomy R\&D, Sandia National Laboratories\thanksref{footnoteinfo}, Albuquerque, NM 87123, USA (e-mail: jslight@sandia.gov).}

\begin{abstract}                
Handling objects with unknown or changing masses is a common challenge in robotics, often leading to errors or instability if the control system cannot adapt in real-time. In this paper, we present a novel approach that enables a six-degrees-of-freedom robotic manipulator to reliably follow waypoints while automatically estimating and compensating for unknown payload weight. Our method integrates an admittance control framework with a mass estimator, allowing the robot to dynamically update an excitation force to compensate for the payload mass. This strategy mitigates end-effector sagging and preserves stability when handling objects of unknown weights. We experimentally validated our approach in a challenging pick-and-place task on a shelf with a crossbar, improved accuracy in reaching waypoints and compliant motion compared to a baseline admittance-control scheme. By safely accommodating unknown payloads, our work enhances flexibility in robotic automation and represents a significant step forward in adaptive control for uncertain environments.
\end{abstract}


\end{frontmatter}

\maketitle
\thispagestyle{empty}
\pagestyle{empty}

\section{INTRODUCTION} 

Robotic manipulation is crucial in various applications including industrial, medical, and field robotics \cite{9663525}. Precise motion and smooth force regulation in dynamic environments remains challenging, aligning with Industry 4.0 goals for adaptive robotic systems \citep{javaid2021substantial}. Admittance control, which enables a robot to modify its motion in response to external forces, has become a useful method \citep{keemink2018admittance, sun2022adaptive}. Distinguishing between different types of forces, such as contacts, unknown payload weights, or sudden impacts, is critical and remains an open challenge in robotic manipulation \citep{hogan1985impedance,
pagliara2024safe}. Recent model-based methods also aim to enhance robustness under uncertainty \citep{slightam2018sliding, slightam2019sliding}.

Other compliant control strategies, such as impedance control, have been widely used for stable and compliant robotic interaction \citep{hogan1985impedance}. However, impedance control methods require actively controlling torques or forces, which can prove to be more challenging to implement on some robotic arm platforms \citep{rojas2022adaptive, ahmed2019robust}. Compliance control approaches in general struggle with real-time adaptation to significant environmental disturbances and payload variations, limiting reliability during dynamic tasks. Thus, there is a need for adaptive methods capable of accurately managing payload variations or external disturbances \citep{zhou2018bcl, nasiri2024teleoperation}.
Since robotic tasks become increasingly complex, real-time compensation for disturbances has become important \citep{su2020internet}. Recent methods address disturbances via sensor-integrated admittance control \citep{haddadin2017robot}, adaptive damping \citep{rojas2022adaptive}, fuzzy logic \citep{li2021fuzzy}, passivity theory, and deep learning adaptive damping \citep{slightam2021passivity, slightam2023deep}.

Controlling a robotic arm in unpredictable environments is a challenging task. Even minor unknown payloads can cause significant deviations from desired trajectories. While admittance control permits compliant deviations for safety, traditional controllers (e.g., PID-based methods) focus strictly on trajectory tracking without inherent compliance, potentially requiring overly conservative strategies or additional sensors to detect disturbances \citep{siciliano2016springer}. For instance, if the payload weight suddenly increases, the robot might experience large positioning errors because it cannot properly separate payload forces from other small disturbances or adapt to changing conditions over time. As a result, the robot may take longer to correct its position, making it less reliable in industrial tasks. 

Despite recent advancements, significant limitations remain in handling unknown payloads during compliant manipulation. To address these issues, our approach integrates real-time mass estimation into an adaptive admittance controller, leveraging force-torque (FT) sensor data to dynamically compensate for unknown payloads. Unlike previous methods, we propose a unified adaptive framework, ensuring compliance and stability simultaneously.

\begin{figure}[ht]
    \centering
    \includegraphics[width=1\linewidth]{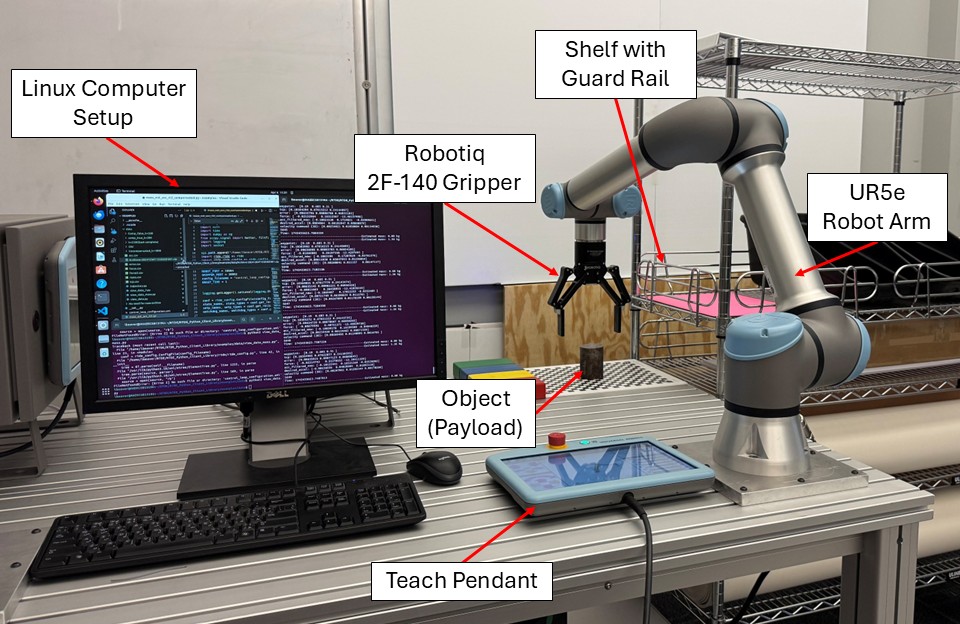}
    \caption{UR5e robot with gripper and shelf setup used in the pick-and-place task.}
    \label{fig:UR5e_setup}
\end{figure}

Our exemplar problem in this manuscript is a bin-to-shelf pick-and place task in an unstructured environment, as shown in Figure~\ref{fig:UR5e_setup}. This application warrants compliance control to not damage items in the unstructured environment. However, manipulating heavier objects with an unknown mass can result in significant deviations from a desired reference trajectory and result in failed pick and place tasks. To solve this challenge, the manuscript presents the design of the admittance controller for a 6-DOF manipulator followed by Section \ref{Sec: solution} that describes the mass adaptive admittance controller. Experimental results and the discussion and conclusions are described in sections \ref{sec:Results} and \ref{sec:conclusion}, respectively.

\section{Admittance Control Design} \label{sec:Formulation}

This section introduces the admittance control approach for a 6-DOF robotic manipulator, which directly uses the FT sensor feedback at the end-effector. Admittance control methods are often more easily implemented than impedance control techniques due to the ability to utilize existing off the shelf motion controllers. 



\begin{figure}[ht]
    \centering
    \begin{tikzpicture}[
    block/.style={draw,align=center,minimum width=2cm,minimum height=1cm}
    ]
    \node[block] (A) at (0,0) {Admittance\\Controller};
    \node[block] (R) at (4,0) {Robot};
    \node[block] (M) at (3.5,1.5) {Environment};
    \node[block] (O) at (0.5,1.5) {-1};
    \draw[->] (A) -- node[above] {$\dot{\bm{p}}$}(R);
    \draw[->] (R) -- ++(1.5, 0) -- ++(0,1.5) -- (M);
    \draw[->] (M) -- node[below] {$F_{ext}$} (O);
    \draw[->] (O) -- node[below] {$F/T$} ++(-2, 0) -- ++(0,-1.5) -- (A);
    \end{tikzpicture}

    \caption{Block diagram for our admittance controller, which sets the reference speed for the manipulator's built-in tracking controller.}
    \label{fig:admittance-flowchart}
\end{figure}
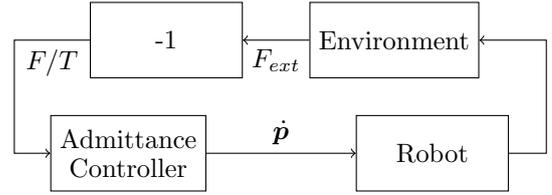

An \emph{admittance controller} in tool space, calculates the desired end-effector velocities based on measured external forces from a FT sensor. The robot’s built-in motion controller then executes these velocity commands each cycle. Figure~\ref{fig:admittance-flowchart} shows an the block diagram of the admittance control architecture, with the robot, interacting with the environment in a closed loop.

An admittance controller is designed to achieve the desired second-order dynamics with a virtual mass, spring, and damper, giving
\begin{equation} \label{eq:linear}
    \bm{F}_{\mathrm{ext}}(t) = M_{a} (\ddot{\bm{p}}_a-\ddot{\bm{p}}_0) + B_{a}(\dot{\bm{p}}_a-\dot{\bm{p}}_0) + K_{a} (\bm{p}_a-\bm{p}_0),
\end{equation}
where $\bm{F}_{\mathrm{ext}}(t) \in \mathbb{R}^3$ is the external force measured at the end effector. 
The matrices $M_{a}$, $B_{a}$, and $K_{a} \in \mathbb{R}^{3\times3}$ are \emph{virtual} mass, damping, and stiffness parameters, selected to achieve the desired admittance behavior and are diagonal, with entries corresponding to the scalar mass, damping, and stiffness constants of an ideal second-order system. 
Finally, $\bm{p}_a$ is the position of the virtual system with ideal admittance, and $\bm{p}_0$ is a reference position.

Taking the Laplace transform of \eqref{eq:linear} with zero initial conditions and taking the derivative yields the impedance form of the dynamical system \citep{ogata2020modern},
\begin{equation} \label{eq:impedance}
   \bm{Z}(s) =  \frac{\bm{F}_{\mathrm{ext}}(s)}{\bm{V}(s)} = \Bigl(M_a s + B_a + \tfrac{K_a}{s}\Bigr)
\end{equation}
which relates the external force $ \bm{F}_{\mathrm{ext}}(s)$ to the velocity $ \bm{V}(s)$.
Inverting \eqref{eq:impedance} results in the admittance:
\begin{equation}
    \begin{aligned}
     \bm{Y}(s) &= \frac{1}{\bm{Z}(s)}= \frac{s}{M_a s^2 + B_a s + K_a},
    \end{aligned}
\end{equation}
where $\bm{Y}(s)$ is the admittance of the ideal second order system.
In a continuous-time, this enables us to generate a command velocity for the end effector.

To reach a static pose, we solve \eqref{eq:linear} for the acceleration, which yields,
\begin{equation}\label{eq:admittance-time}
    \ddot{\bm{p}}_{\mathrm{a}} = M_a^{-1}\left(\bm{F}_{ext} -B_a \dot{\bm{p}}_{\mathrm{a}} - K_a( \bm{p}_a -\bm{p}_0) \right).
\end{equation}
Eq. \eqref{eq:admittance-time} describes the ideal admittance acceleration $\ddot{\bm{p}}_{\mathrm{a}}$ from our admittance model while considering a static reference pose $\bm{p}_0$, the ideal velocity $\dot{\bm{p}}_{\mathrm{a}}$, and external forces $\bm{F_{ext}}$.

The admittance velocity is obtained by integrating the admittance acceleration, determined by
\begin{equation} \label{eq:velocity_integral_simplified}
\dot{\bm{p}}_{\mathrm{a}}(t)
=
\int_{t_0}^{t} \ddot{\bm{p}}_{\mathrm{a}}(t) \, dt.
\end{equation}

The admittance velocity can be commanded directly to the robot's onboard controller. 
Implementation of \eqref{eq:velocity_integral_simplified} enables real-time compliance, where the robot naturally yields or ``admits'' to external forces while following the reference trajectory.

\section{Mass Adaptive Admittance Control} \label{Sec: solution}
\subsection{Adaptive Admittance Formulation} \label{Subsec: adaptive admittance}
\begin{figure}[h!]
    \centering
    \includegraphics[width=0.85\linewidth]{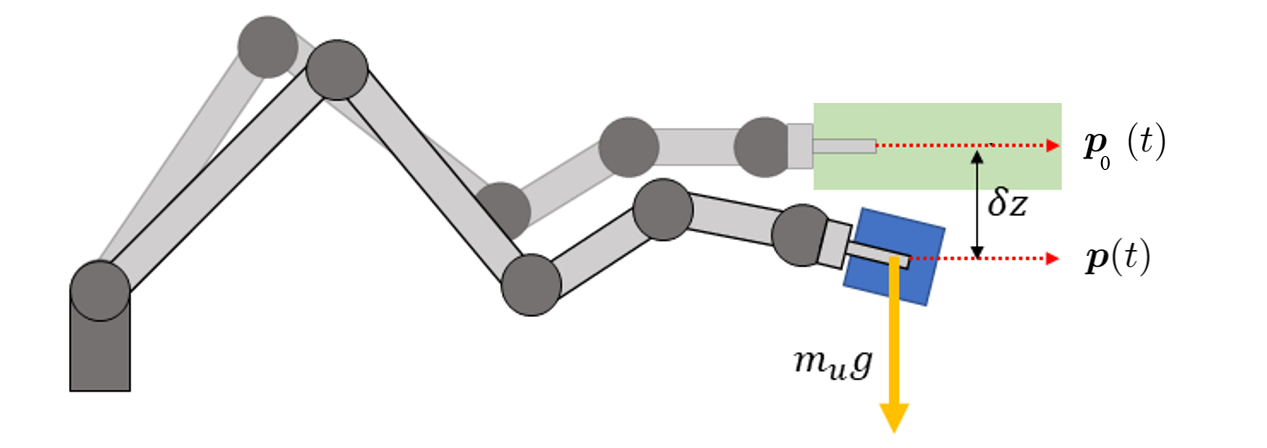}
    \caption{Unknown payload mass $m_u$ causing sag by an amount $\delta_z$ from the original desired pose $p_0$.}
    \label{fig:admittance_problem}
\end{figure}

One significant drawback of admittance control are the large deviations that can occur while following the reference trajectory when picking and placing an object of significant unknown mass. To compensate for this, an accurate model of the mass properties of an unknown object must be attained.
This drawback is illustrated in Fig. \ref{fig:admittance_problem}, which shows the resulting deviation of a admittance controller when moving a mass of an unknown object. 

This displacement, denoted as \(\delta_z\), is influenced by the virtual stiffness of the system and can be expressed as
\begin{equation} \label{eq: sagging}
\delta_z = \frac{m g}{K_{zz}},
\end{equation}

where \(m\) is the mass of the payload, \(g\) is the gravitational acceleration, and \(K_{zz}\) is the virtual stiffness in the \(z\)-direction. Sagging from lifting an unknown payload introduces a vertical offset, $\delta _z$, in the end-effector's position, which highlights the need to compensate the controller to better track the reference trajectory in free space.

To improve the system's reference  tracking while manipulating an object, and to achieve our objective of putting an item on a shelf without colliding with the environment, the unknown forces from the added mass are compensated for by using a virtual \textit{excitation} force, \( F_{\text{exc}} \) as an additional control input, allowing the system to adjust its response dynamically. Considering \eqref{eq:linear}, the resulting admittance dynamics, incorporating \( F_{\text{exc}} \), are formulated as \citep{mcarthur2024coordinated},
\begin{equation}
\begin{aligned}
\label{eq:admittance}
\bm{F}_{\text{ext}} 
= 
M_a \bigl(\ddot{\bm{p}}_{\mathrm{a}} -\ddot{\bm{p}}_{\mathrm{0}} \bigr)
\;+\;
B_a \bigl(\dot{\bm{p}}_{\mathrm{a}} - \dot{\bm{p}}_{\mathrm{0}}\bigr) \\
\;+\;
K_a \bigl(\bm{p}_a - \bm{p}_{0}\bigr) 
\;-\;
\bm{F}_{\text{exc}}.
\end{aligned}
\end{equation}
To compensate for unknown payload mass, we upload the external force $F_{\text{ext}}$ after performing mass estimation during free space trajectory following. The forces due to the payload can subsequently be compensated for with the excitation force, $F_{exc}$ to eliminate sagging.


\vspace{0.2cm}
\subsection{Mass Estimation and Excitation Force} \label{Subsec: mass estimate}

Considering Newton's second law we have
\begin{equation} \label{eq:newton}
    (m_g+m_u)\ddot{\bm{p}} = (m_g + m_u)\bm{g} + \bm{f}(t),
\end{equation}
where \(m_u + m_g\) is the total mass at the end effector, including the known gripper mass \(m_g\), the unknown payload mass \(m_u\), \(\ddot{\bm{p}}\) as the measured acceleration of the manipulator’s end-effector, \(\bm{g} \) is gravitational acceleration vector, and \(\bm{f}(t)\) is the reaction force measured by the FT sensor.
To compensate for the unknown payload, we subtract off the gravitational contribution by projecting both \(\bm{f}\) and \(\ddot{\bm{p}}\) onto the vertical axis via a dot product with the unit vector \(\hat{\bm{z}}\),
\begin{equation} \label{eq:mass-dot}
    m_u 
    \;=\; 
    \frac{\bm{f} \cdot \hat{\bm{z}}}
          {\ddot{\bm{p}} \cdot \hat{\bm{z}} \;-\; \bm{g} \cdot \hat{\bm{z}}}
    \;-\; 
    m_g.
\end{equation}

For simplicity we define
\[
  f_{z} (t) = \bm{f}(t) \cdot \hat{\bm{z}},  \quad
  \ddot{p}_{z}(t) = \left(\ddot{\bm{p}}(t) - \bm{g}\right) \cdot \hat{\bm{z}},
\]

which simplifies \eqref{eq:mass-dot} to
\begin{equation} \label{eq:mass final}
    m_u 
    \;=\;
    \frac{f_z(t)}{\ddot{p}_z(t)}
    \;-\; 
    m_g.
\end{equation}

By updating \(m_u\) in real time, the total mass \(m = m_g + m_u\) is provided to the admittance controller to compensate for gravity, reducing end-effector sag by injecting the updated excitation force in \eqref{eq:admittance} as 
\begin{equation}\label{eq:excitation_force}
    \bm{F}_{exc} = m_u (\ddot{p}_z)\hat{\bm{z}},
\end{equation}

which helps attain better reference trajectory tracking while maintaining compliance for additional disturbances.


\subsection{Stability Analysis for On-line Adaptation} \label{Subsec:stability}
\begin{remark} \label{rmk:stability}
    In free space, a mass estimator with transfer function $\hat{M}_u(s)$ is stable if and only if the polynomial,
    \begin{equation}
        \Big(M_as^2 +B_as+K_a\Big)\Big(M_u-\hat{M}_u(s)\Big)s+R(s) = 0
    \end{equation}
    has roots with a negative real part, where $R(s)$ is the closed-loop transfer function of the robot arm.
\end{remark}

\begin{proof}
    The robot arm experiences environment forces in a feedback loop as depicted in Fig. \ref{fig:admittance-flowchart}. 
    The forward path transfer function is,
    \begin{equation}
        Y_{ol}(s) = \frac{s\,R(s)}{M_a s^2 + B_as + K_a}.
    \end{equation}
    In free space, the environmental force acting on our force torque sensor is a pure mass,
    \begin{equation}
        F_{ext}(s) = \frac{1}{(M_u - \hat{M}_u)s^2},
    \end{equation}
    where $M_u$ is the unknown mass we estimate with $\hat{M}_u(s)$. 
    The closed-loop transfer function for the system is,
    \begin{equation*}
        Y(s) = 
        \frac{s^2\,R(s)\,\big(M_u - \hat{M}_u(s)\big)}
        {\Big(M_as^2 +B_as+K_a\Big)\Big(M_u - \hat{M}_u(s)\Big)s + R(s)}.
    \end{equation*}
    This system is bounded-input bounded-output stable if and only if all poles have roots with negative real parts.
\end{proof}

\section{Experimental Results} \label{sec:experiment}

We implemented our experiments on a UR5e robot arm equipped with an FT sensor mounted on the robot wrist, internal accelerometer, and a Robotiq 2F-140 gripper. A Linux PC communicates with the robot using Real-Time Data Exchange (RTDE), handling admittance control, mass estimation, and gripper operations based on joint angles, end-effector pose, FT measurements, and accelerometer data.

Because FT sensor and accelerometer data are originally in the sensor frame, we transformed these measurements into a common inertial (base) frame using Rodrigues' rotation formula \citep{murray2017mathematical}. After transforming, we calculated \(\bm{f}^b_{\text{comp}}(t)\) by subtracting sensor offsets, known gripper weight, and gravity components, which isolates the net external force from unknown payloads. We then applied a moving average filter to obtain smoothed force \(\bm{f}^b_{\text{filtered}}(t)\) and acceleration \(\ddot{\bm{p}}^b_{\text{filtered}}(t)\), which were fed into the admittance controller \eqref{eq:admittance-time} and mass estimator \eqref{eq:mass final}.

We implement \eqref{eq:admittance-time} using scalar mass, damping, and stiffness values, applied uniformly across all Cartesian axes. We set $m$ to a nominal value that approximates the combined mass of the gripper and the expected payload. Next, we select the damping coefficient $b$ based on the critical-damping relationship:
\begin{equation} \label{eq:damper}
b = 2 \sqrt{m \cdot k}, \quad\text{(critical damping)}
\end{equation}
to reduce overshoot. The stiffness $k$ is chosen to achieve the desired stiffness. All parameters were empirically selected on the basis of the system performance during the task.

Using filtered acceleration and force measurements, we computed the admittance acceleration \(\ddot{\bm{p}}_{\text{adm}}\) and integrated it numerically to get desired end-effector velocity, which subsequently we send as commands to the robot's internal velocity controller.

We specified a series of waypoint positions \(\{\bm{p}_{0,i}\}\) in the base frame to define the task. At runtime, the robot moves toward the current waypoint until the Euclidean position error falls below a defined threshold $\epsilon$. Once this condition is met, the robot proceeds to the next waypoint.

Table~\ref{tab:params} summarizes the main parameters used in our experiments. The position-error threshold $\epsilon$ was set to 3.5\,mm, and stiffness $k$ was varied between 100 and 2500\,N/m to evaluate different compliance conditions.

\begin{table}[h!]
    \centering
    \caption{Summary of Key Parameters}
    \label{tab:params}
    \begin{tabular}{c c c}
    \hline
    \textbf{Parameter} & \textbf{Value/Range} & \textbf{Eq.} \\
    \hline
    $m$        & 4\,kg                & --       \\
    $b$        & $2\sqrt{m \cdot k}$ & \eqref{eq:damper} \\
    $k$        & 100--2500\,N/m      & --       \\
    $\epsilon$ & 3.5\,mm             & -- \\
    \hline
    \end{tabular}
\end{table}

\section{Experimental Setup and Results} \label{sec:Results}

We conducted four experiments to evaluate how different stiffness settings and mass compensation affect the robot’s ability to complete a bin-to-shelf pick-and-place task while the admittance controller is implemented.

\begin{figure}[ht]
    \centering
    \includegraphics[width=0.8\linewidth]{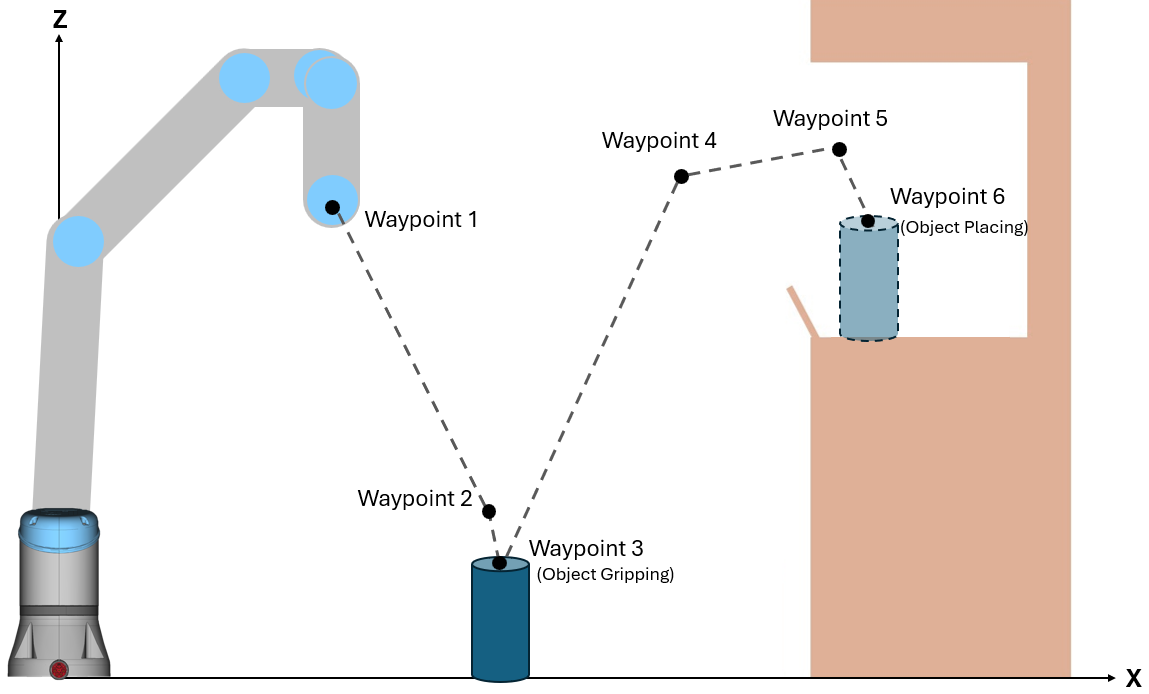}
    \caption{Schematic of the six planned waypoints on the X--Z plane. Waypoint 3 is used for
    object grasping, and waypoint 6 is where the object is placed on the shelf.}
    \label{fig:Schematic_object_PP}
\end{figure}

Figure~\ref{fig:Schematic_object_PP} provides a schematic overview of the planned pick-and-place
trajectory in the X--Z plane, illustrating how the robot moves through six waypoints, from
the pick location to the final place position on the shelf.

Table~\ref{tab:overview} summarizes the main parameters and outcomes of each experiment.
We vary the virtual stiffness $k$ from low to high values and enable or disable real-time mass compensation. In all trials, the arm moved a 1500 \,g payload through the same set of six waypoints.

\begin{table}[ht]
    \centering
    \caption{Summary of Experiments}
    \label{tab:overview}
    \begin{tabular}{lcccc}
    \hline
    \textbf{Exp.} & \textbf{$K$ }& \textbf{$M$ Comp.} & \textbf{Completion} & \textbf{Sag (mm)} \\
    \hline
    1 & Medium & No  & Fail & 8.1 \\
    2 & High & No  & \textbf{Success} & 3.5 \\
    3 & Low  & Yes & \textbf{Success} & $\bm{<3.5}$ \\
    4 & Low  & No  & Fail & 46.8 \\
    \hline
    \end{tabular}
\end{table}

\emph{Experiment 1) Medium Stiffness, No Compensation}

\begin{figure}[ht] 
    \centering \includegraphics[width=0.85\linewidth]{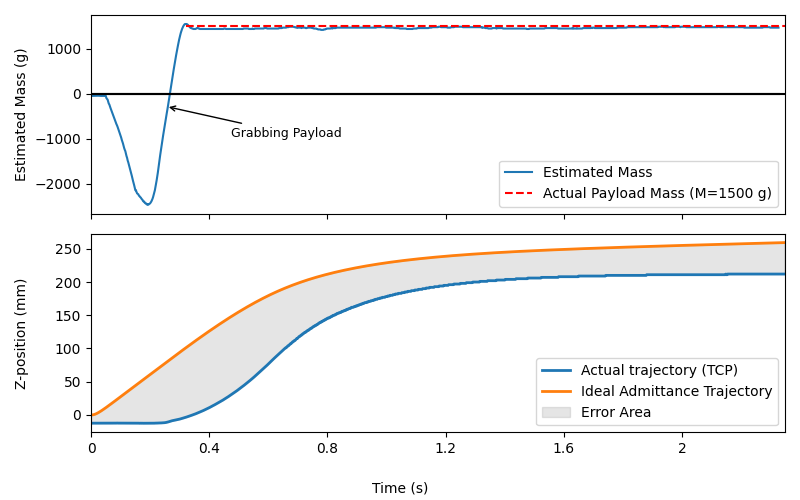} 
    \caption{Mass estimation (top) and actual vs. ideal admittance trajectories in the Z-axis without compensation (bottom).} \label{fig:medium_k} 
\end{figure}

In the first experiment, we set the stiffness to $k=1800\,\mathrm{N/m}$, without applying mass compensation. This stiffness alone does not fully overcome the payload’s weight in the $z$-direction, causing a persistent sag. Due to uncompensated sagging, the end-effector cannot maintain its expected elevation, which can be seen in the trajectory deviation shown in Figure~\ref{fig:medium_k}; it also demonstrates the performance of our mass estimator. Before grasping the payload, the mass estimate is approximately zero.
After grasping, around $t \approx 0$s in the figure, the arm sags 
slightly downward in the vertical ($z$) direction 
and comes into contact with the table. 
That causes a \emph{negative} mass estimate due to the reaction force. Once the gripper lifts the object and there is no longer contact between the table, 
the estimate stabilizes on average around $1460\,\mathrm{g} \pm 30\,\mathrm{g}$ at $t\approx0.4$s after grasping, reflecting minor fluctuations from sensor inaccuracy or drift.
This demonstrates 
the robot's ability to estimate the mass in real-time, with an approximate $\pm30\,\mathrm{g}$ variation due to sensor inaccuracy or drift. According to the manufacturer’s datasheet, the sensor’s accuracy is approximately 407\,g (4\,N).

The bottom plot compares the actual end-effector trajectory in the Z-direction with the ideal response generated by the admittance model. The shaded area highlights the positional discrepancy caused by uncompensated payload forces. This confirms that \(k=1800\,\mathrm{N/m}\) is insufficient to compensate for the payload weight without an additional mass compensation strategy.

\emph{Experiment 2) High Stiffness, No Compensation}

Our second experiment uses a \emph{higher} virtual stiffness ($k=2500\,\mathrm{N/m}$)
. Although this approach does not involve mass compensation, by increasing stiffness it is sufficient to overcome the payload's weight and the sagging. The top plot again indicates a momentary negative mass estimation when the end‐effector contacts the object located on the table, but once the gripper lifts the object, the robot maintains a Z‐axis offset below the threshold. Consequently, the manipulator advances through all six waypoints, showing that raising stiffness can avoid large sag and achieve task completion. However, this will cause the robot and implemented admittance controller would not be compliant to other external forces. Increasing \(k\) would cause sag reduction but also reduce compliance/admittance to external forces. Instead, once our mass‐estimation process compensates for the payload (demonstrated in the next experiments), the end‐effector tracks the waypoints successfully \emph{without} imposing unnecessarily large stiffness.

\emph{Experiment 3) Low Stiffness with Mass Compensation}

In the third experiment, we implemented \emph{true} mass compensation by inputting our estimated mass data directly into our admittance controller, which is calculated by the explained and validated (by the first and the second experiments) estimated mass method. 
By tuning \(k\) properly, we prevent the robot from becoming too stiff, ensuring it remains gentle enough to respond to external forces and disturbances appropriately while still adapting to disturbances or changes in unmodeled payloads. 

\begin{figure} [ht]
    \centering
    \includegraphics[width=1 \linewidth]{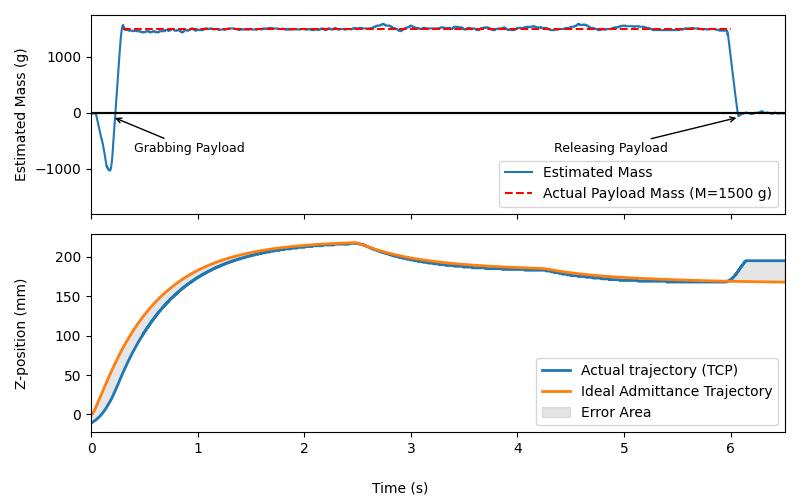}
    \caption{Mass estimation (top) and actual vs. ideal admittance trajectories in the Z-axis (bottom).}
    \label{fig:Compensated_mass}
\end{figure}

    Figure~\ref{fig:Compensated_mass} shows the performance of the third experiment with mass compensation. The shaded error area between the actual and ideal admittance trajectories confirms that, once the payload is grasped and the mass is estimated, the end-effector closely adapts to the reference path and successfully completes the pick-and-place task.

\emph{Experiment 4) Low Stiffness, No Compensation}

Finally, we repeated the low-stiffness test but disabled mass compensation to illustrate the sagging effect. In this experiment, as it was expected, after picking the object, the manipulator failed to achieve the next waypoint, confirming that mass estimation is essential for handling larger payloads at low stiffness. The results are closely similar to Figure~\ref{fig:medium_k} (\textit{Experiment 1, Medium Stiffness}), but with a significantly larger error of  46.8 mm due to the lower stiffness. We have omitted the redundant figure and result visualization for brevity.

To quantify how well the end-effector maintains vertical accuracy during the pick-and-place task, we compute the Root Mean Square Error (RMSE) in the z-direction over the critical portion of the trajectory after payload grasping. Specifically, we use the deviation between the admittance controller’s integrated position output $\bm{p}_a$ and the reference pose $\bm{p}_0$ in the vertical direction. The RMSE is defined as:

\begin{equation} \label{eq:EMSE}
\mathrm{RMSE} = \sqrt{\frac{1}{N}\sum_{i=1}^N \|\bm{p}_{a,z}(i) - \bm{p}_{0,z}(i)\|^2}.
\end{equation}
The RMSE remains low before the payload is picked up, as no significant external force is applied. However, after grasping, RMSE increases significantly in conditions without mass compensation, demonstrating the resulting sag. Table~\ref{tab:rmse} summarizes the RMSE values observed across all four experimental conditions.

\begin{table}[ht]
    \centering
    \caption{RMSE of Position Tracking in Each Experiment}
    \label{tab:rmse}
    \begin{tabular}{lc}
    \hline
    \textbf{Experiment} & \textbf{RMSE [mm]} \\
    \hline
    1: $k=1800$, No Comp. & 9.530 \\
    2: $k=2500$, No Comp. & 4.705 \\
    3: $k=300$,  + Mass Comp. & $\bm{1.988}$ \\
    4: $k=300$,  No Comp. & 20.584 \\
    \hline
    \end{tabular}
\end{table}

Table~\ref{tab:rmse} shows the RMSE of the end-effector's position for each experiment. Since experiment~3  has the lowest RMSE, it confirms that the mass compensation effectively offsets unknown payloads without sacrificing compliance. By contrast, Experiment~4 has the highest RMSE, showing that low stiffness alone leads to significant sagging and inaccuracy. 

\begin{figure} [ht]
    \centering
    \includegraphics[width=1 \linewidth]{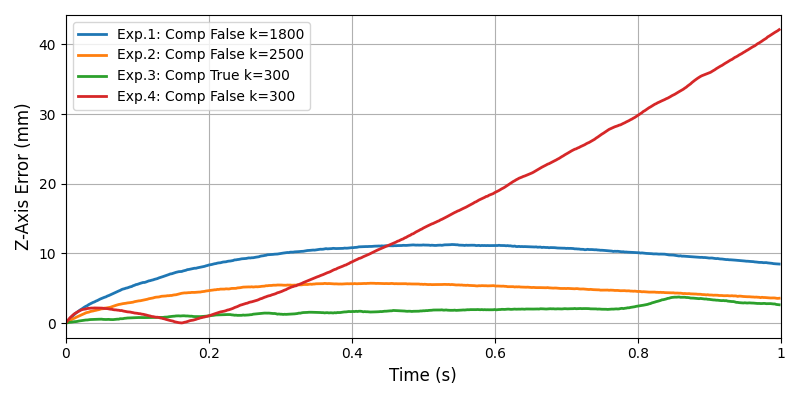}
    \caption{Z-axis position error over a critical segment of 500 time steps after payload grasping, comparing performance across four different stiffness and compensation conditions.}
    \label{fig:errorz_experiments}
\end{figure}

Figure~\ref{fig:errorz_experiments} further illustrates the instantaneous Z-axis position error ($p_{\text{adm,z}} - p_{\text{d,z}}$) across the critical segment. Experiment 3, employing adaptive mass compensation, achieves the lowest error, clearly demonstrating its effectiveness in minimizing sagging. In contrast, Experiment 4, with low stiffness and no compensation, exhibits substantial error growth, emphasizing the necessity of adaptive mass compensation in low-stiffness scenarios.

Overall, the results show that increasing stiffness helps reduce sagging but also makes the system less responsive to external forces. A stiffness of \(K=300\) without compensation led to too much sagging, while increasing it to \(K=2500\) reduced the sagging but made the system less flexible. Our proposed approach, which includes real-time mass estimation, successfully compensates for unknown payloads while keeping the system both stable and flexible. This allows the end-effector to complete the pick-and-place task accurately without needing very high stiffness, making it better at handling external disturbances.

\section{Conclusions} \label{sec:conclusion}
In this work, we presented an adaptive admittance controller with real-time payload mass estimation to handle unknown loads in a 6-DOF robotic manipulator. Our experiments on a UR5e platform show that online mass estimation effectively reduces end-effector sagging, keeping the vertical offset below the desired threshold once the payload mass is compensated. Though demonstrated on the vertical axis, our method can extend naturally to multiple axes and rotations.

We also showed that simply increasing virtual stiffness can offset unknown payloads, but it comes at the expense of decreased compliance and lower tolerance to small external forces. In contrast, integrating mass compensation allows the manipulator to “push back” and respond more gently against unexpected contacts improving both safe interaction and accurate waypoint tracking as a key goal in different demands like Industry~4.0 environments.

Future work will extend adaptive admittance and mass compensation to full 6-DOF control, integrating additional sensor fusion and advanced impedance strategies \citep{Hossein2024development}. 

\bibliography{bibs/references,bibs/beaver_pubs}

\end{document}